\pdfoutput=1
\documentclass[11pt]{article}

\usepackage{emnlp2021}

\usepackage{times}
\usepackage{latexsym}
\usepackage{enumitem}
\setlist[itemize]{noitemsep}

\usepackage[T1]{fontenc}
\usepackage[utf8]{inputenc}

\usepackage{microtype}

\usepackage{booktabs}

\title{Evaluation of Abstractive Summarisation Models with Machine Translation in Deliberative Processes}

 \author{Miguel Arana-Catania\textsuperscript{1,2}, Rob Procter\textsuperscript{1,2}, Yulan He\textsuperscript{1,2}, Maria Liakata\textsuperscript{2,3} \\
         \textsuperscript{1}Department of Computer Science, University of Warwick, UK \\ \textsuperscript{2}Alan Turing Institute, UK \\ \textsuperscript{3}Queen-Mary University of London, UK}

\begin{document}
\maketitle
%\vspace{-1.0cm}
\begin{abstract}
We present work on summarising deliberative processes for non-English languages. Unlike commonly studied datasets, such as news articles, this deliberation dataset reflects difficulties of combining multiple narratives, mostly of poor grammatical quality, in a single text. We report an extensive evaluation of a wide range of abstractive summarisation models in combination with an off-the-shelf machine translation model. Texts are translated into English, summarised, and translated back to the original language. We obtain promising results regarding the fluency, consistency and relevance of the summaries produced. Our approach is easy to implement for many languages for production purposes by simply changing the translation model.
\end{abstract}

\section{Introduction}

The processes of deliberation and collective intelligence production have evolved radically thanks to the possibility of carrying them out digitally. However, this often results in large amounts of generated content in the deliberations, causing information overload that prevents their potential from being fully realised \cite{arana2021citizen, davies2020online, davies2021mixed}. To address this, we evaluate the potential value of abstractive summarisation models when combined together with a machine translation system in synthesising and filtering information collected through such processes. Whereas the current technology of language models is mostly limited to a few languages, which creates a barrier to their more widespread use, our approach can be deployed for many languages just by changing the translation model without the need to generate new, ad-hoc corpora for the task or costly retraining for each new language. The current evaluation is done in a Spanish deliberation dataset.

We have carried out an evaluation with 6 abstractive summarisation models: BART \citep{lewis2019bart}, T5 \citep{raffel2019exploring}, BERT (PreSumm – BertSumExtAbs: \citealp{liu2019text}), PG (Pointer-Generator with Coverage Penalty) \citep{see2017get}, CopyTransformer \citep{gehrmann2018bottom}, and FastAbsRL \citep{chen2018fast}. Those models are applied in combination with the machine translation system MarianMT \citep{junczys2018marian} using the Opus-MT models \citep{tiedemann2020opus}. We have evaluated the quality of the summaries for each model and their comparison.

Early research on the problem of text summarisation in low resourced languages (although not focused on deliberation)~\citet{orasan2008evaluation} demonstrated the limitations of machine translation systems at that time. Recently, \citet{ouyang2019robust} revisited the problem of low quality translations in low resourced languages and successfully demonstrated the possibility of using abstractive summarisation by retraining their model on corpora that have gone through the same machine translation process. In this study, we complete the cycle, translating from the original language to English, summarising, and translating back to the original language, thus avoiding the need for retraining.

Using other approaches, \citet{yao2015phrase} studied English-to-Chinese summarisation combining an extractive approach with a process of sentence compression that effectively abstracts the results. \citet{duan2019zero}, following \citet{shen2018zero}, exploited the capability of a resource-rich language summariser in a teacher-student framework that connects it to the target language summariser.

\section{Dataset}
The evaluation has been carried out with a dataset from deliberative processes in Spanish, which was translated into English to carry out the summarisation. The generated summaries were then translated back into Spanish for evaluation. Thus, the evaluators evaluated summaries of Spanish texts.

The dataset is available in the Madrid City Council ‘Datos Abiertos’ repository\footnote{\url{https://datos.madrid.es}}, called `Comentarios'. It contains public deliberations in relation to citizen proposals submitted to the participation platform of the city council. The dataset has been selected due to the great success of the participation platform, which has led to $26,400$ proposals and $125,135$ comments being submitted. This is one of the most successful cases of digital participation in the world and is therefore a perfect case study for evaluating the information overload problem in deliberative processes \cite{arana2021citizen}.

Each proposal presents a debate space where public comments can be found. Forty debates were selected covering different deliberation scenarios in the dataset. These represent three cases: $20$ debates with ($n = 10$) comments, the most common case of debates with few comments; $15$ debates with ($20 \leq n \leq 30$) comments, for the medium case; and $5$ debates with ($60 \leq n \leq 70$) comments, the large number of comments case.

The debates were also selected to cover three different comment scenarios, i.e., from very short to very lengthy comments. In the first scenario from $1,000$ to $5,000$ total characters; in the medium scenario from $3,000$ to $13,000$; and in the large scenario from $10,000$ to $18,000$ characters. For each debate the text to summarise was created by concatenating its comments into a single text.

By using debates from all scenarios regarding the number of comments and comment length we ensure that the selection is not biased to a specific scenario of deliberation that could skew our results.
Examples of the debates can be found in the \nameref{appendix}, illustrating the combination of multiple narratives through the different comments and the poor grammatical quality of the texts.

\section{Abstractive Summarisation Methodology}
Different models were selected, covering some of the best available summarisers, but also different model architectures:

\begin{itemize}
    \item BART \citep{lewis2019bart}\footnote{\label{huggface}Implementation by HuggingFace \url{https://github.com/huggingface/transformers}}. This combines a bidirectional transformer as an encoder, similar to the following T5 and BERT cases, with a left-to-right autoregressive decoder similar as GPT \citep{radford2018improving}. The ‘large-cnn’ pre-trained model\textsuperscript{\ref{huggface}} has been used here.
    \item T5 \citep{raffel2019exploring}\textsuperscript{\ref{huggface}}. This uses an encoder-decoder transformer architecture, trained in the Colossal Clean Crawled Corpus. The ‘t5-small’ pre-trained model\textsuperscript{\ref{huggface}} has been used.
    \item BERT (PreSumm – BertSumExtAbs: \citealp{liu2019text})\footnote{\label{presumm}Implementation by the authors \url{https://github.com/nlpyang/PreSumm}}. This uses a BERT \citep{devlin2018bert} encoder and a randomly initialized transformer as a decoder, fine-tuning it first as an extractive summariser and then as an abstractive one. The BertSumExtAbs pre-trained model\textsuperscript{\ref{presumm}} has been used.
    \item PG (Pointer-Generator with Coverage Penalty) \citep{see2017get}\footnote{\label{openmt}Implementation by OpenNMT \url{https://opennmt.net/OpenNMT-py/examples/Summarization.html}}. This uses a 1-layer bidirectional LSTM encoder and a 1-layer unidirectional LSTM decoder with attention, with the possibility of switching between copying words or generating them (Pointer-Generator) and including a coverage mechanism adding up attention distributions of previous steps to minimise repetitions. The 'OpenNMT BRNN (2 layer, emb 256, hid 1024)' pre-trained model\textsuperscript{\ref{openmt}} has been used.
    \item CopyTransformer \citep{gehrmann2018bottom}\footnote{OpenNMT implementation thanks to \url{https://github.com/sebastianGehrmann/bottom-up-summary}}. This uses the transformer architecture, but one attention head defines the copy distribution. The 'OpenNMT Transformer' pre-trained model\textsuperscript{\ref{openmt}} has been used.
    \item FastAbsRL \citep{chen2018fast}\footnote{Implementation by the authors \url{https://github.com/ChenRocks/fast_abs_rl}}. An extractor agent is used to select sentences (using LSTM layers to represent and copy sentences) and an abstractor network is used to compress and paraphrase the selected sentences. Both are trained separately and then the full model is trained with reinforcement learning by using A2C \citep{mnih2016asynchronous}.
\end{itemize}

The reported Rouge scores of these models \citep{lin2004rouge} are shown in Table \ref{tab:rouge}. None of the pre-trained models used were retrained.

\begin{table}
\centering{\small
\begin{tabular}{ l c c c}
\toprule
\textbf{Model} & \textbf{R1} & \textbf{R2} & \textbf{RL} \\
\midrule
BART - large-cnn & 44.16 & 21.28 & 40.90 \\
T5 - t5-small & 41.12 & 19.56 & 38.35 \\
BERT - BertSumExtAbs & 42.13 & 19.60 & 39.18 \\
PG - OpenNMT – BRNN  & 39.12 & 17.35 & 36.12 \\
CopyT - OpenNMT & 39.25 & 17.54 & 36.45 \\
FastAbsRL & 40.88 & 17.80 & 38.54 \\
\bottomrule
\end{tabular}}
\caption{Rouge scores reported on the CNN/DailyMail dataset \citep{hermann2015teaching}.}
\label{tab:rouge}
\vspace{-6mm}
\end{table}

Additional models were also evaluated: Adversarial Reinforce GAN \citep{wang2018learning}, using Generative Adversarial Networks; Contextual Matching \citep{zhou2019simple}, joining ELMo with a domain fluency model; PoDA \citep{wang2019denoising}, denoising autoencoder transformer with a pointer-generator layer; and GenParse \citep{song2018structure}, combining sequential word generation with tree-based parsing. Our initial qualitative evaluation found that none of them were competitive enough with the selected models. Several of these models work at the sentence level, which may impact their relevance in our deliberative case, where texts are composed of multiple authors' comments.

The machine translation system used was MarianMT \citep{junczys2018marian} using its HuggingFace implementation, with Opus-MT models\footnote{\url{https://github.com/Helsinki-NLP/Opus-MT}} developed by the Helsinki-NLP group.

Machine translation was first applied to the original text of the deliberations before applying the summarisers, and then to the summary generated to convert back to the original language (see \nameref{appendix}). Thus, even when the summarisation models are trained with English datasets, the full system can be used in deliberations of any language supported by the machine translation system. The Opus-MT models used in this work count currently with pre-trained models for 1738 language pairs. It is left for future work to evaluate the effect of the translation model, and to apply it to other languages to determine their quality. The models used here show a good performance (see BLEU scores in \citeauthor{OpusMTen,OpusMTes}) for the languages used.

\section{Evaluation Design}
We developed a protocol for the human evaluation of the summaries generated by the different models, following designs used in previous studies (\citealp{amplayo2020unsupervised}; \citealp{liu2019text}; \citealp{narayan2018ranking}; \citealp{paulus2017deep}; \citealp{yoon2020learning}; \citealp{song2018structure}). First, the different models were compared regarding their relative overall quality using the Best-Worst scaling \citep{louviere2015best}, shown to be more accurate than a generic individual scoring model, and simultaneously reducing the number of assessments required \citep{kiritchenko2017best}.

For each debate, $6$ different summaries were generated, one for each of the models to be evaluated. These summaries were organised in $9$ tuples of $4$ elements each, where each summary appeared in $6$ of the tuples in random order not allowing the evaluator to identify each model used. In total, considering all the debates, $360$ tuples were produced. Each of these tuples was evaluated by $5$ independent evaluators (native Spanish speakers with a minimum education level of a Bachelor’s degree), producing a total of $1,800$ evaluations. The score for each summary consisted of the percentage of times it was evaluated as Best, minus the percentage of times it was evaluated as Worst.

In addition, a second evaluation was carried out for two summaries in each debate. The models were selected randomly in each case, while ensuring that each model had the same number of evaluations. Here, we were interested in whether the models produce results of sufficient quality to be useful to participants in the debate. Thus, we we used an absolute rather than a relative score. We asked evaluators to rate the following (definitions were shared with evaluators) on a Likert scale from 1 (Strongly disagree) to 4 (Strongly agree):

\begin{itemize}
    \item Informativeness/Relevance. The summary contains the most relevant ideas and positions of the debate.
    \item Fluency/Readability/Grammaticality. The summary sentences are grammatically correct, easy to read and understand (considering as a baseline the fluency of the original debate).
    \item Consistency/Faithfulness. The ideas or facts contained in the summary appear in the original debate.
    \item Creativity. The summary has been written with its own words and sentences (instead of copying sentences directly from the debate).
\end{itemize}

\section{Evaluation Results}
The results obtained for the overall comparison between models are shown in Table \ref{tab:comp}, which reports the average scores of all the evaluators.

\begin{table}
\centering\small{
\begin{tabular}{l c c c c}
\toprule
\textbf{Model} & \textbf{comp} & $\sigma$ & \textbf{comp\textsubscript{[0,100]}} & $\sigma$\textsubscript{[0,100]} \\
\midrule
BART & 33.08 & 11 & 66.54 & 5 \\
BERT & 23.33 & 10 & 61.67 & 5 \\
PG & 6.25 & 13 & 53.13 & 6 \\
T5 & -16.08 & 14 & 41.96 & 7 \\
CopyT & -16.42 & 5 & 41.79 & 2 \\
FastAbsRL & -30.17 & 10 & 34.92 & 5 \\
\bottomrule
\end{tabular}}
\caption{Comparison scores using the Best-Worst scaling (and thus in the range $[-100,100]$) with its standard deviation, and normalised to the $[0,100]$ range.}
\label{tab:comp}
\vspace{-2mm}
\end{table}

Paired Student’s $t$-tests were performed between all pairs of models to confirm that the difference was statistically significant. This is not the case for the BERT and BART models ($p=0.09$), showing very close results. There is also a clear overlap between T5 and CopyTransformer. All the other combination pairs are found to have a statistically significant difference ($p < 0.05$).

These results are in line with the previous results on English datasets that BART and BERT are the top two summarisers \cite{lewis2019bart,liu2019text}. However, in the present evaluation the performance of a state-of-the-art model (T5) falls below that of a much older model (PG).

\begin{table}
\centering{\small
\begin{tabular}{l c c c c}
\toprule
\textbf{Model} & \textbf{Informative} & $\sigma$ & \textbf{Fluent} & $\sigma$ \\
\midrule
BART & 2.58 & 0.8 & 2.85 & 0.8  \\
BERT & 2.53 & 0.8 & 2.65 & 0.9  \\
PG & 2.33 & 0.7 & 2.28 & 0.8  \\
T5 & 2.50 & 0.8 & 2.30 & 0.8  \\
CopyT & 2.14 & 0.6 & 2.02 & 0.8  \\
FastAbsRL & 2.02 & 0.7 & 1.73 & 0.6  \\
\midrule
& \textbf{Consistent} & $\sigma$ & \textbf{Creative} & $\sigma$ \\
\midrule
BART & 2.88 & 0.8 & 2.08 & 0.7 \\
BERT & 2.72 & 0.9 & 1.98 & 0.6 \\
PG & 2.67 & 0.8 & 2.02 & 0.6 \\
T5 & 2.63 & 0.9 & 1.97 & 0.6 \\
CopyT & 2.46 & 0.9 & 1.81 & 0.6 \\
FastAbsRL & 2.13 & 0.7 & 1.82 & 0.7 \\
\bottomrule
\end{tabular}}
\caption{Rating and SD for each model on a scale from 1 (Strongly disagree) to 4 (Strongly agree).}
\label{tab:rating}
\vspace{-6mm}
\end{table}

The results for the evaluation of the qualitative aspects of each summariser are shown in Table \ref{tab:rating}. It is important to note that in this case the standard deviation is larger compared to the first case, which is due to the smaller number of evaluations, and thus the following comments should take into account their statistical significance. 

In this individual evaluation of each model, it can be seen again how BART obtains the best ratings in all four categories evaluated. BERT is the second best for the categories of ‘\emph{Informativeness}’, ‘\emph{Fluency}’ and ‘\emph{Consistency}’, while PG jumps to the second position for ‘\emph{Creativity}’. T5 is in the third position for the categories ‘\emph{Informativeness}’ and ‘\emph{Fluency}’ and PG is the third best for ‘\emph{Consistency}’.

This confirms the best results of BART and BERT, and a close result for T5 for generating informative summaries, but a poorer result for fluency. This may be the reason why the T5 model performed worse in the general overall comparison.

BART and BERT perform well in terms of `\emph{consistency}', with scores close to 3. They perform a bit worse for `\emph{fluency}' and `\emph{informativeness}', around the middle of the possible rating 2.5. Regarding `\emph{creativity}', the models have a poor performance, with a score of around 2, meaning that they tend to copy instead of paraphrase.

\section{Conclusions}
In this study we have evaluated the application of state-of-the-art, abstractive summarisation models to deliberative processes in Spanish using an off-the-shelf machine translation model. Although we focused on Spanish in this study, our proposed pipeline can be easily deployed without additional effort to many other languages. This offers significant benefits for production applications (especially cases dealing with wide ranges of languages) that are rarely available in other approaches that usually need to be tuned for each language. However, the evaluation of the quality for other languages is left for future work. We have done a comparative evaluation of the overall quality of the models, and an evaluation of each model with respect to different qualitative aspects: \emph{informativeness, fluency, consistency}, and \emph{creativity}.

As a general conclusion, from the models evaluated BART and BERT produced the best results, and satisfactory results are obtained in the proposed pipeline for the quality of the summaries. With regard to the most important aspects, the models show a good result for the categories of \emph{fluency} and \emph{consistency}, and an average result regarding the \emph{informativeness}. These results are especially promising considering the complexity and low grammatical fluency and consistency involved in texts typical of deliberative processes. BART and BERT are the only models that score above the middle score in each of the three categories, and thus we argue perform sufficiently well to be used in practice.

\section*{Acknowledgements}
This work was funded in part by the UK Engineering and Physical Sciences Research Council (grant no. EP/V048597/1). RP is supported by a Turing Fellowship (grant no. EP/N510129/1). YH and ML are supported by Turing AI Fellowships (grant no. EP/V020579/1, EP/V030302/1, respectively).

\bibliography{anthology,custom}

\begin{thebibliography}{33}
\expandafter\ifx\csname natexlab\endcsname\relax\def\natexlab#1{#1}\fi

\bibitem[{Amplayo and Lapata(2020)}]{amplayo2020unsupervised}
Reinald~Kim Amplayo and Mirella Lapata. 2020.
\newblock Unsupervised opinion summarization with noising and denoising.
\newblock \emph{arXiv preprint arXiv:2004.10150}.

\bibitem[{Arana-Catania et~al.(2021)Arana-Catania, Lier, Procter, Tkachenko,
  He, Zubiaga, and Liakata}]{arana2021citizen}
Miguel Arana-Catania, Felix-Anselm~Van Lier, Rob Procter, Nataliya Tkachenko,
  Yulan He, Arkaitz Zubiaga, and Maria Liakata. 2021.
\newblock Citizen participation and machine learning for a better democracy.
\newblock \emph{Digital Government: Research and Practice}, 2(3):1--22.

\bibitem[{Chen and Bansal(2018)}]{chen2018fast}
Yen-Chun Chen and Mohit Bansal. 2018.
\newblock Fast abstractive summarization with reinforce-selected sentence
  rewriting.
\newblock \emph{arXiv preprint arXiv:1805.11080}.

\bibitem[{Davies et~al.(2021)Davies, Arana-Catania, Procter, van Lier, and
  He}]{davies2021mixed}
Jonathan Davies, Miguel Arana-Catania, Rob Procter, Felix-Anselm van Lier, and
  Yulan He. 2021.
\newblock Evaluating the application of nlp tools on mainstream participatory
  budgeting processes in scotland.
\newblock In \emph{Proceedings of the International Conference on Theory and
  Practice of Electronic Governance}, pages 317--320.

\bibitem[{Davies and Procter(2020)}]{davies2020online}
Jonathan Davies and Rob Procter. 2020.
\newblock Online platforms of public participation: a deliberative democracy or
  a delusion?
\newblock In \emph{Proceedings of the 13th International Conference on Theory
  and Practice of Electronic Governance}, pages 746--753.

\bibitem[{Devlin et~al.(2018)Devlin, Chang, Lee, and
  Toutanova}]{devlin2018bert}
Jacob Devlin, Ming-Wei Chang, Kenton Lee, and Kristina Toutanova. 2018.
\newblock Bert: Pre-training of deep bidirectional transformers for language
  understanding.
\newblock \emph{arXiv preprint arXiv:1810.04805}.

\bibitem[{Duan et~al.(2019)Duan, Yin, Zhang, Chen, and Luo}]{duan2019zero}
Xiangyu Duan, Mingming Yin, Min Zhang, Boxing Chen, and Weihua Luo. 2019.
\newblock Zero-shot cross-lingual abstractive sentence summarization through
  teaching generation and attention.
\newblock In \emph{Proceedings of the 57th Annual Meeting of the Association
  for Computational Linguistics}, pages 3162--3172.

\bibitem[{Gehrmann et~al.(2018)Gehrmann, Deng, and Rush}]{gehrmann2018bottom}
Sebastian Gehrmann, Yuntian Deng, and Alexander~M Rush. 2018.
\newblock Bottom-up abstractive summarization.
\newblock \emph{arXiv preprint arXiv:1808.10792}.

\bibitem[{Hermann et~al.(2015)Hermann, Ko{\v{c}}isk{\`y}, Grefenstette,
  Espeholt, Kay, Suleyman, and Blunsom}]{hermann2015teaching}
Karl~Moritz Hermann, Tom{\'a}{\v{s}} Ko{\v{c}}isk{\`y}, Edward Grefenstette,
  Lasse Espeholt, Will Kay, Mustafa Suleyman, and Phil Blunsom. 2015.
\newblock Teaching machines to read and comprehend.
\newblock \emph{arXiv preprint arXiv:1506.03340}.

\bibitem[{Junczys-Dowmunt et~al.(2018)Junczys-Dowmunt, Grundkiewicz, Dwojak,
  Hoang, Heafield, Neckermann, Seide, Germann, Aji, Bogoychev
  et~al.}]{junczys2018marian}
Marcin Junczys-Dowmunt, Roman Grundkiewicz, Tomasz Dwojak, Hieu Hoang, Kenneth
  Heafield, Tom Neckermann, Frank Seide, Ulrich Germann, Alham~Fikri Aji,
  Nikolay Bogoychev, et~al. 2018.
\newblock Marian: Fast neural machine translation in c++.
\newblock \emph{arXiv preprint arXiv:1804.00344}.

\bibitem[{Kiritchenko and Mohammad(2017)}]{kiritchenko2017best}
Svetlana Kiritchenko and Saif~M Mohammad. 2017.
\newblock Best-worst scaling more reliable than rating scales: A case study on
  sentiment intensity annotation.
\newblock \emph{arXiv preprint arXiv:1712.01765}.

\bibitem[{Lewis et~al.(2019)Lewis, Liu, Goyal, Ghazvininejad, Mohamed, Levy,
  Stoyanov, and Zettlemoyer}]{lewis2019bart}
Mike Lewis, Yinhan Liu, Naman Goyal, Marjan Ghazvininejad, Abdelrahman Mohamed,
  Omer Levy, Ves Stoyanov, and Luke Zettlemoyer. 2019.
\newblock Bart: Denoising sequence-to-sequence pre-training for natural
  language generation, translation, and comprehension.
\newblock \emph{arXiv preprint arXiv:1910.13461}.

\bibitem[{Lin(2004)}]{lin2004rouge}
Chin-Yew Lin. 2004.
\newblock Rouge: A package for automatic evaluation of summaries.
\newblock In \emph{Text summarization branches out}, pages 74--81.

\bibitem[{Liu and Lapata(2019)}]{liu2019text}
Yang Liu and Mirella Lapata. 2019.
\newblock Text summarization with pretrained encoders.
\newblock \emph{arXiv preprint arXiv:1908.08345}.

\bibitem[{Louviere et~al.(2015)Louviere, Flynn, and Marley}]{louviere2015best}
Jordan~J Louviere, Terry~N Flynn, and Anthony Alfred~John Marley. 2015.
\newblock \emph{Best-worst scaling: Theory, methods and applications}.
\newblock Cambridge University Press.

\bibitem[{Mnih et~al.(2016)Mnih, Badia, Mirza, Graves, Lillicrap, Harley,
  Silver, and Kavukcuoglu}]{mnih2016asynchronous}
Volodymyr Mnih, Adria~Puigdomenech Badia, Mehdi Mirza, Alex Graves, Timothy
  Lillicrap, Tim Harley, David Silver, and Koray Kavukcuoglu. 2016.
\newblock Asynchronous methods for deep reinforcement learning.
\newblock In \emph{International conference on machine learning}, pages
  1928--1937. PMLR.

\bibitem[{Narayan et~al.(2018)Narayan, Cohen, and Lapata}]{narayan2018ranking}
Shashi Narayan, Shay~B Cohen, and Mirella Lapata. 2018.
\newblock Ranking sentences for extractive summarization with reinforcement
  learning.
\newblock \emph{arXiv preprint arXiv:1802.08636}.

\bibitem[{OpusMTen(2020)}]{OpusMTen}
OpusMTen. 2020.
\newblock \href {https://huggingface.co/Helsinki-NLP/opus-mt-en-es}
  {https://huggingface.co/helsinki-nlp/opus-mt-en-es}.

\bibitem[{OpusMTes(2020)}]{OpusMTes}
OpusMTes. 2020.
\newblock \href {https://huggingface.co/Helsinki-NLP/opus-mt-es-en}
  {https://huggingface.co/helsinki-nlp/opus-mt-es-en}.

\bibitem[{Orǎsan and Chiorean(2008)}]{orasan2008evaluation}
Constantin Orǎsan and Oana~Andreea Chiorean. 2008.
\newblock Evaluation of a cross-lingual romanian-english multi-document
  summariser.
\newblock \emph{Proceedings of the 6th International Conference on Language
  Resources and Evaluation, LREC 2008}.

\bibitem[{Ouyang et~al.(2019)Ouyang, Song, and McKeown}]{ouyang2019robust}
Jessica Ouyang, Boya Song, and Kathleen McKeown. 2019.
\newblock A robust abstractive system for cross-lingual summarization.
\newblock In \emph{Proceedings of the 2019 Conference of the North American
  Chapter of the Association for Computational Linguistics: Human Language
  Technologies, Volume 1 (Long and Short Papers)}, pages 2025--2031.

\bibitem[{Paulus et~al.(2017)Paulus, Xiong, and Socher}]{paulus2017deep}
Romain Paulus, Caiming Xiong, and Richard Socher. 2017.
\newblock A deep reinforced model for abstractive summarization.
\newblock \emph{arXiv preprint arXiv:1705.04304}.

\bibitem[{Radford et~al.(2018)Radford, Narasimhan, Salimans, and
  Sutskever}]{radford2018improving}
Alec Radford, Karthik Narasimhan, Tim Salimans, and Ilya Sutskever. 2018.
\newblock Improving language understanding by generative pre-training (2018).

\bibitem[{Raffel et~al.(2019)Raffel, Shazeer, Roberts, Lee, Narang, Matena,
  Zhou, Li, and Liu}]{raffel2019exploring}
Colin Raffel, Noam Shazeer, Adam Roberts, Katherine Lee, Sharan Narang, Michael
  Matena, Yanqi Zhou, Wei Li, and Peter~J Liu. 2019.
\newblock Exploring the limits of transfer learning with a unified text-to-text
  transformer.
\newblock \emph{arXiv preprint arXiv:1910.10683}.

\bibitem[{See et~al.(2017)See, Liu, and Manning}]{see2017get}
Abigail See, Peter~J Liu, and Christopher~D Manning. 2017.
\newblock Get to the point: Summarization with pointer-generator networks.
\newblock \emph{arXiv preprint arXiv:1704.04368}.

\bibitem[{Shen et~al.(2018)Shen, Chen, Yang, Liu, Sun et~al.}]{shen2018zero}
Shi-qi Shen, Yun Chen, Cheng Yang, Zhi-yuan Liu, Mao-song Sun, et~al. 2018.
\newblock Zero-shot cross-lingual neural headline generation.
\newblock \emph{IEEE/ACM Transactions on Audio, Speech, and Language
  Processing}, 26(12):2319--2327.

\bibitem[{Song et~al.(2018)Song, Zhao, and Liu}]{song2018structure}
Kaiqiang Song, Lin Zhao, and Fei Liu. 2018.
\newblock Structure-infused copy mechanisms for abstractive summarization.
\newblock \emph{arXiv preprint arXiv:1806.05658}.

\bibitem[{Tiedemann and Thottingal(2020)}]{tiedemann2020opus}
J{\"o}rg Tiedemann and Santhosh Thottingal. 2020.
\newblock Opus-mt--building open translation services for the world.
\newblock In \emph{Proceedings of the 22nd Annual Conference of the European
  Association for Machine Translation}, pages 479--480.

\bibitem[{Wang et~al.(2019)Wang, Zhao, Jia, Li, and Liu}]{wang2019denoising}
Liang Wang, Wei Zhao, Ruoyu Jia, Sujian Li, and Jingming Liu. 2019.
\newblock Denoising based sequence-to-sequence pre-training for text
  generation.
\newblock \emph{arXiv preprint arXiv:1908.08206}.

\bibitem[{Wang and Lee(2018)}]{wang2018learning}
Yau-Shian Wang and Hung-Yi Lee. 2018.
\newblock Learning to encode text as human-readable summaries using generative
  adversarial networks.
\newblock \emph{arXiv preprint arXiv:1810.02851}.

\bibitem[{Yao et~al.(2015)Yao, Wan, and Xiao}]{yao2015phrase}
Jin-ge Yao, Xiaojun Wan, and Jianguo Xiao. 2015.
\newblock Phrase-based compressive cross-language summarization.
\newblock In \emph{Proceedings of the 2015 conference on empirical methods in
  natural language processing}, pages 118--127.

\bibitem[{Yoon et~al.(2020)Yoon, Yeo, Jeong, Yi, and Kang}]{yoon2020learning}
Wonjin Yoon, Yoon~Sun Yeo, Minbyul Jeong, Bong-Jun Yi, and Jaewoo Kang. 2020.
\newblock Learning by semantic similarity makes abstractive summarization
  better.
\newblock \emph{arXiv preprint arXiv:2002.07767}.

\bibitem[{Zhou and Rush(2019)}]{zhou2019simple}
Jiawei Zhou and Alexander~M Rush. 2019.
\newblock Simple unsupervised summarization by contextual matching.
\newblock \emph{arXiv preprint arXiv:1907.13337}.

\end{thebibliography}
\bibliographystyle{acl_natbib}

\clearpage

\appendix

\section{Appendix}\label{appendix}
We present below an example of a debate used in the evaluation in Spanish and its machine translation to English. Following them we present the summaries generated using T5, FastAbsRL, BART, and BERT. Finally, we include the translations of these summaries.

The texts are presented in the same order used in the project. We start with a debate in Spanish, which is translated into English. This translated debate is summarised, and finally the summary is translated back into Spanish. The evaluators analysed only the original debate in Spanish and the final summaries in Spanish.

\subsection{Original Spanish debate}
\begin{itemize}
    \item ademas proponemos tranvía.
    \item el casco no es obligatorio para mayores de 15 años mientras circulan en ciudad. lo dice la dgt.por lo demás, te doy la razón. deben cumplir la normativa de circulación. pero, eh!... los conductores de coches y motos también. hay demasiados que no respetan a los ciclistas... ¿sabias que en ciudad, un ciclista debe ocupar 1 carril de circulación... y no ir por el borde?.
    \item se deberían sancionar las bicis que van por las aceras o fuera de los carriles bicis.
    \item si las bicis van por las aceras es porque es muy peligroso ir por los carriles de los coches aunque estén marcados. no existe concienciación todavía por parte de los usuarios conductores. por otro lado, el hecho en sí de ir por la acera no es peligroso, siempre que se vaya "a paso de peatón". lo que no se puede es ir rápido.para mí el verdadero peligro es en las horas nocturnas, en que muchos ciclistas van sin luz alguna y no se ven hasta que estás prácticamente encima de ellos... eso en amsterdam está rigurosamente prohibido y se multa. aquí he visto a la policía municipal pasar de todo al verlos....
    \item obviamente quien dice eso no ha cogido una bici en su vida, el casco en bici no salva vidas, es un hecho, salva vidas el conductor respetuoso.
    \item nunca,pero nunca jamás he visto parar un ciclista en un semáforo rojo,o se suben a la acera para cruzar sorteando a los peatones o directamente se lo saltan,en un paso de peatones menos se paran.¿qué pasa,que las norma no son para todos por igual? si un coche se salta un semáforo,la multa es bestial! un poco más de respeto,sobre todo cuando circulan por la acera a la velocidad que les da la gana,con el peligro que conlleva.se creen que todo vale y la calle es suya.
    \item se puede circular por la calzada, aunque haya carril bici vecin@.
    \item no me lo creo....nunda digas nunca!.
    \item ¿no cree que está generalizando demasiado? no todos van con auriculares, no todos se saltan los semáforos, y los coches se tienen que aconstumbrar a la presencia de las bicis....es un medio de transporte más, y se merece respeto.
    \item la obligación del casco desincentiva el uso d ela bicicleta, que en el caso de mardid está mejorando la movilidad sin aumentar la contaminación
\end{itemize}

\subsection{Machine translated debate}
\begin{itemize}
    \item and we're proposing a tram.
    \item the helmet is not mandatory for more than 15 years as they travel in the city. says dgt. otherwise, I give you the reason. they must comply with the traffic regulations. but, uh!... the drivers of cars and motorcycles also. there are too many that do not respect cyclists... did you know that in the town, a cyclist must occupy 1 lane of traffic... and not go by the edge?.
    \item bikes that go along the sidewalks or off the bike lanes should be sanctioned.
    \item if the bikes go through the sidewalks it is because it is very dangerous to go down the lanes of the cars even if they are marked. there is no awareness yet on the part of the users drivers. On the other hand, the fact itself of going down the sidewalk is not dangerous, as long as it goes "by foot".What you can not do is go fast.For me the real danger is in the night hours, in which many cyclists go without any light and you don't see until you are practically above them... that in Amsterdam is rigorously forbidden and is fined. here I have seen the municipal police pass everything when you see them....
    \item obviously whoever says that hasn't taken a bike in his life, the bike helmet doesn't save lives, it's a fact, it saves lives the respectful driver.
    \item never, but I've never seen a cyclist stop at a red light, or get on the sidewalk to cross by shooting pedestrians or directly jump him, at a pace of pedestrians less stop.What happens, that the rules are not for everyone alike? if a car jumps a light, the ticket is best! a little more respect, especially when they circulate along the sidewalk at the speed that gives them the desire, with the danger that it carries. they believe that everything is good and the street is theirs.
    \item you can drive along the road, even if there is a nearby bicycle lane.
    \item I don't believe it... never say it!
    \item Don't you think it's generalizing too much? Not everyone goes with headphones, not everyone jumps the traffic lights, and cars have to settle for the presence of bikes....it's one more means of transport, and it deserves respect.
    \item the obligation of the helmet to discourage the use of bicycles, which in the case of Tuesday is improving mobility without increasing pollution
\end{itemize}

\subsection{Generated summaries}
\begin{itemize}
\item \textbf{T5}. the rules are not for everyone alike. not everyone jumps the traffic lights, not everyone goes with headphones, and not everybody jumps traffic lights. a little more respect, especially when they circulate along the street at the speed that gives them the desire, with the danger that it carries. I don't believe it... never say it! Don't you think it's generalizing too much? Don'
    \item \textbf{FastAbsRL}. the helmet is not mandatory for more than 15 years .''
the real danger is in which many cyclists go without any light and you don't see until you are practically above them... that in amsterdam is rigorously forbidden .
otherwise, i give you the reason. .
they must comply with the traffic 
    
    \item \textbf{BART}. Bikes that go along the sidewalks or off the bike lanes should be sanctioned. The real danger is in the night hours, in which many cyclists go without any light. Not everyone goes with headphones, not everyone jumps the traffic lights, and cars have to settle for the presence of bikes. It's one more means of transport, and it deserves respect. 
    \item \textbf{BERT}. the helmet is not mandatory for more than 15 years as they travel in the city
it is because it is very dangerous to go down the lanes of the cars even if they are marked
there is no awareness yet on the part of the users drivers drivers 
\end{itemize}

\subsection{Machine translated summaries}
\begin{itemize}
    \item \textbf{T5}. las reglas no son para todos por igual. no todos saltan los semáforos, no todos van con auriculares, y no todo el mundo salta semáforos. un poco más de respeto, especialmente cuando circulan por la calle a la velocidad que les da el deseo, con el peligro que conlleva. No lo creo... nunca lo digo! ¿No crees que está generalizando demasiado?
    \item \textbf{FastAbsRL}. el casco no es obligatorio durante más de 15 años. "
el verdadero peligro es en el que muchos ciclistas van sin ninguna luz y no se ve hasta que usted está prácticamente por encima de ellos... que en amsterdam está rigurosamente prohibido.
Si no, te doy la razón.
deben cumplir con el tráfico.
    \item \textbf{BART}. Las bicicletas que van por las aceras o fuera de los carriles bici deben ser sancionadas. El verdadero peligro es en las horas de la noche, en las que muchos ciclistas van sin ninguna luz. No todos van con auriculares, no todos saltan los semáforos, y los coches tienen que conformarse con la presencia de bicicletas. Es un medio de transporte más, y merece respeto.
    \item \textbf{BERT}. el casco no es obligatorio por más de 15 años ya que viajan por la ciudad
es porque es muy peligroso ir por los carriles de los coches, incluso si están marcados
todavía no hay conciencia por parte de los conductores de los usuarios
\end{itemize}

\end{document}